\documentclass[letterpaper, 10 pt, conference]{ieeeconf} 
\usepackage[utf8]{inputenc}
\usepackage{todonotes}
\usepackage{amsmath}
\usepackage{caption}
\usepackage{subcaption}

\makeatletter
\def\bstctlcite{\@ifnextchar[{\@bstctlcite}{\@bstctlcite[@auxout]}}
\def\@bstctlcite[#1]#2{\@bsphack
  \@for\@citeb:=#2\do{%
    \edef\@citeb{\expandafter\@firstofone\@citeb}%
    \if@filesw\immediate\write\csname #1\endcsname{\string\citation{\@citeb}}\fi}%
  \@esphack}
\makeatother 

\title{ANSEL Photobot: A Robot Event Photographer with Semantic Intelligence}
\author{Dmitriy Rivkin, Gregory Dudek, Nikhil Kakodkar, \\
David Meger, Oliver Limoyo, Michael Jenkin, Xue Liu, \\
Francois Hogan}
\date{September 2022}

\begin{document}
\bstctlcite{BSTcontrol}

\maketitle

\section{Abstract}

Our work examines the way in which large language models can be used for robotic planning and sampling, specifically the context of automated photographic documentation.  Specifically, we illustrate how  to produce a photo-taking robot with an exceptional level of semantic awareness by leveraging recent advances in general purpose language (LM) and vision-language (VLM) models. Given a high-level description of an event we use an LM to generate a natural-language list of photo descriptions that one would expect a photographer to capture at the event. We then use a VLM to identify the best matches to these descriptions in the robot's video stream. The photo portfolios generated by our method are consistently rated as more appropriate to the event by human evaluators than those generated by existing methods.

\section{Introduction}
This paper presents ANSEL (Appropriate sNap SELection) Photobot, the world's first semantically-aware robot photographer that can take photos across multiple domains starting only with high-level English task descriptions. It is implemented using publicly available language and language-vision models with no fine-tuning. Event photographers are expected to obey social conventions in their photos, and which photos are appropriate to take are highly dependent on the nature event and the activities that are likely to occur. For example, at the christening of a large ship, a key activity was historically the breaking of a champagne bottle on the bow of the ship (and, in fact, special extra-fragile bottles are available for this purpose).  
The ability to have robots act on these conventions has only been 
unlocked in the past couple of years due to advances in general purpose 
LMs~\cite{gpt3,lambda,palm} 
and VLMs~\cite{clip, flamingo}. 
LMs contain much world knowledge and can perform common sense reasoning but operate in a abstract language space, while VLMs can ground the language to the robot's sensory reality. Though their application to robotics is in the early stages, they are already beginning to unlock new robotics capabilities in the domains of manipulation \cite{saycan2022arxiv,inner_monologue} and navigation \cite{lm_nav}.

\begin{figure}
\includegraphics[width=0.48\textwidth]{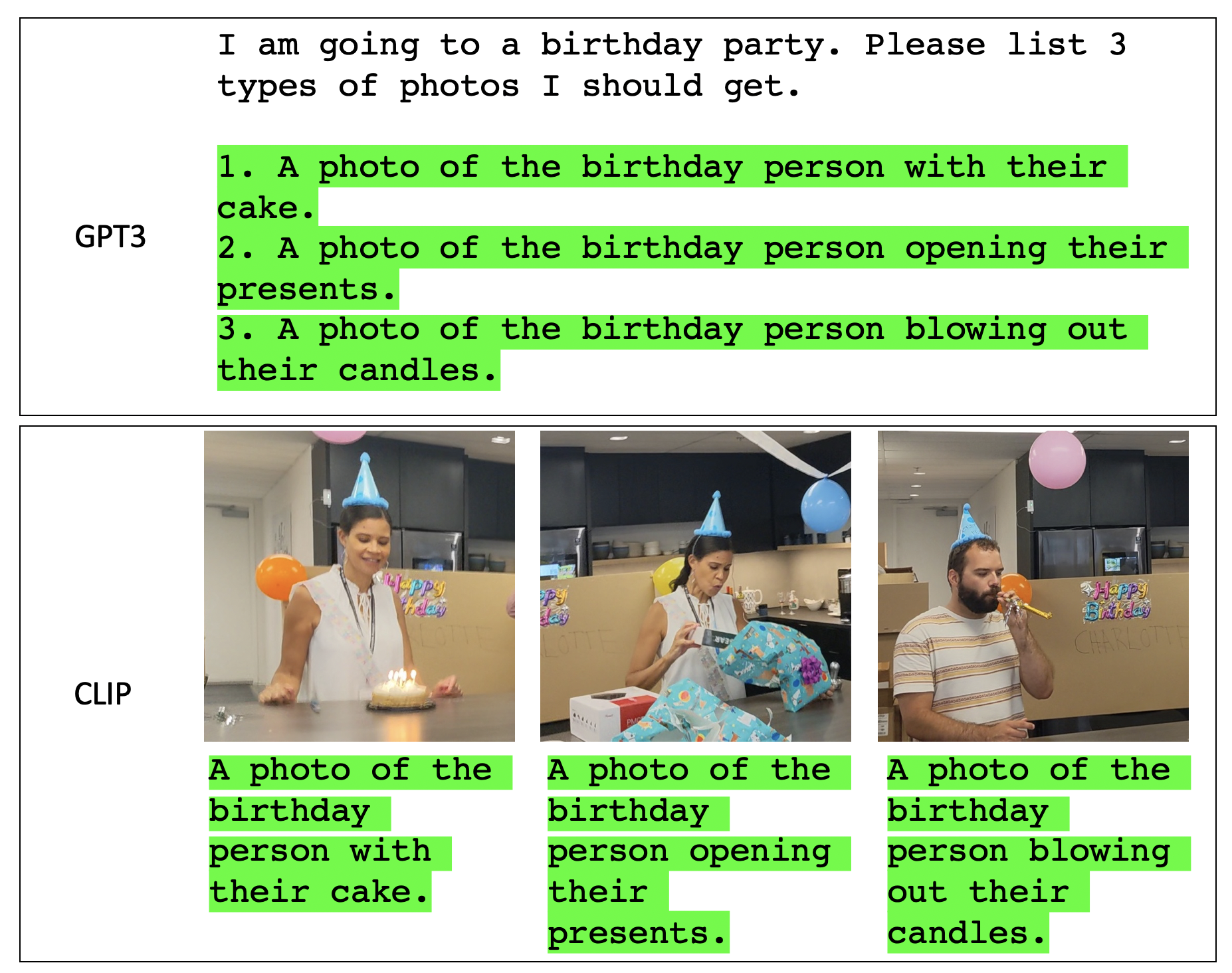}
\centering
\caption{Our method gets good ideas for photos from an LM (GPT3 \cite{gpt3}, top), then finds them in the robot's visual data stream using a VLM (CLIP \cite{clip}, bottom). Green highlighting indicates text generated by GPT3. The mismatch between the rightmost caption and image illustrates some of the challenges involved in this approach, and is discussed further in the text below.}
\label{fig:example}
\end{figure}

 The bulk of the prior work in this domain focuses on exploiting the language models' understanding of the world of places and objects, while our work highlights their capabilities in the domain of social convention. We capture a video stream from a robot present at a party, then extract a fixed number of the most contextually appropriate images from it (Figure \ref{fig:example}). We refer to this set of images as a ``portfolio.'' In order to evaluate our method, we created robot-centric video recordings of three social events, and generated 9-image portfolios of each using both our method and a modern video summarization technique (CA-SUM,  \cite{casum}). We then asked workers to perform pairwise comparisons between portfolios, ranking which they believe is more appropriate given the event description. Our method was consistently rated as more appropriate than those created by CA-SUM.

The primary contributions of this work are:
\begin{enumerate}
    \item We propose an approach to the development of robot event photographers with an unprecedented level of general-purpose semantic awareness.
    \item We describe the design of a functioning system to embody this idea and we evaluate its performance.
    \item We discuss some of the trade-offs and design  considerations.
\end{enumerate} 
Since the design of all aspects of a novel functions system like this is necessarily complex and multi-layered,  we focus here in the novel aspects of the overall design and use of an LM, and space limitations preclude a detailed description of the hardware and embodiment.

\section{Related work} 

\subsection{Robot photographers}
Robot photographers can be used both to collect personal snapshots, or to document various events activities such as an ongoing parade or an environmental
phenomenon~\cite{manderson2017robotic}.  When the  problem includes the more comprehensive motion of documenting some phenomenon, the term ``photo documenter' or ``photo-documentarian'' might be more apt (technically, the term ``documentarian'' also refers to photographer specializing in producing a factual record'').

Until now, work in the field of robot photography has been focused primarily on snapping good photos, where the ``goodness'' of the photos was evaluated individually either by volunteers on a Likert scale as in \cite{robot_photographer_luke, robot_photographer0, robot_photographer1}, or a with an existing photo composition quality model \cite{robot_photographer_lerop}. All of these works present hardware designs as well as the accompanying software.

Earlier work, such as \cite{robot_photographer0} and \cite{robot_photographer_luke}, focus on the tasks of detecting people (e.g. using skin detection), navigating to them, and then composing the photograph using heuristics such as the ``rule of thirds.'' Later works employed more machine-learning focused techniques. For example, \cite{robot_photographer_lerop} used deep reinforcement learning to learn a template-matching policy to capture photos of people in desired poses. \cite{robot_photographer1}.

Finally, and perhaps most similar to our work in terms of experimental setup, Newbury~{\em et al.} considered a robot that could navigate along a pre-programmed trajectory and collect large numbers of photographs of people~\cite{robot_photographer1}. These photos are then analyzed post-hoc to detect the best ones using a learned photo quality predictor. The authors of this paper note that this closely resembles the behavior of real modern-day photographers who have access to large memory cards.

Robot photographers have been considered in several contexts, from those that optimize image quality from an essentially fixed viewpoint, to those that track and follow a subject  of interest. Applications for such robots range from those that might optimize portraiture, to automatically record complex procedures such as surgical procedures or military operations.
Some authors have focused to the local actions of a robot photographer such as framing the shots and observing the rules associated with capturing a good composition~\cite{dixon2003picture}.  A larger-scale class of problems relate to planning the trajectories of one or more cameras 
for visual coverage of some phenomenon~\cite{bourque1998viewpoint}, including maintaining visual content with the subject being recorded~\cite{ShkurtiMaintainingVis2014,ShkurtiPursuit2018},  planning paths that allow an escaping photographic subject to be captured~\cite{ShkurtiDudek2017}, or encircling a subject with a swarm of robot paparrazi~\cite{jenkin2000paparazzi,brisset2006paparazzi,BaldiPaparazziTheory2018}.

\subsection{Video Summarization}
Video summarization is the problem of extracting summaries from videos, either in the form of a set of short clips or a collection of images\cite{ca_sum, generic_summarization0}.  Summarization can  be based on simple methods
like uniform sampling, image content, geometric considerations (where it has some connections to SLAM), and combinations of location, sensor data, and scene content~\cite{flint2010growing,girdhar2016modeling}.

Google Inc released a commercial product called ``Clips'' that was a stationary camera that automatically selected when to take short video sequences based on on-device face recognition (note that Google's AI-enabled Clips product is distinct from the CLIP software system from OpenAI that we use in this paper).  Unfortunately, the lack of viewpoint variability and other pragmatic considerations limited the acceptance of the camera~\cite{bonnington2018google} and thus some innovators have even considered endowing it with mobility~\cite{pierce2020roomba+}.

Query-guided video summarization, first defined in \cite{tvsum}, is similar, except that a user query is also provided. For example, if the video was a tour of a national park, and the query was ``water," then the resultant summary should contain a representative subset of the different scenes of water in the video. Examples include \cite{clip_it, tvsum, query_sum0}.

While the video summarization task, and especially the query guided version, superficially seem quite similar to the robotic photography task as defined in our work, there is a key difference. Namely, our task requires inferring the appropriate photographs to get based on a high-level description, while query-guided image summarization is more object focused. For example, for a video of a birthday party, our task would ask ``get all the pictures one would expect from a birthday party", while query guided video summarization would ask ``get a representative selection of all the pictures of cake."

\section{Preliminaries}
Pretrained LMs and VLMs are key to our method. We consider approaches based on the Transformer~\cite{NIPS2017_3f5ee243} architecture, scaled up to very large model sizes trained on very large amounts of data and available for public use. We access GPT3 \cite{gpt3} through its web interface, and run CLIP \cite{clip} on our own server using publicly available checkpoints. In this section we briefly review each of these models, as well as CA-SUM, a generic video summarization technique which we use as baseline.

\subsection{Transformers}

Encoder and decoder models have emerged as a flexible approach for numerous language learning tasks~\cite{scalingLLM,googleTranslate}, allowing reasoning to occur in a latent embedding space rather than the complex tokens representing raw language. Concomitantly, attention has emerged as a key tool in the recurrent processing of text, allowing long-term correlations to be captured~\cite{badhanauAlignAndTranslate,kim2017structured}. The use of (masked) multi-head self-attention in both the encoder and decoder, known as Tranformers~\cite{NIPS2017_3f5ee243}, combine these advances, allowing attention to summarize information across both temporal scales and levels of abstraction. Transformers provide opportunities for parallelization, reducing training duration by several orders of magnitude while improving performance, which has been a key to scaling to internet-sized datasets, making Transformers the architecture of choice for modern semantic models.

\subsection{GPT3}

The scaling potential of Transformers allows training on language datasets containing billions of tokens. The GPT-2~\cite{radford2019language} approach  demonstrated the power of unsupervised training for multi-task learning of many related language tasks, with a model of 1.5B parameters, achieving unprecedented performance upon publication.  GPT3~\cite{gpt3} further scaled-up GPT-2s architecture to 175B parameters by exploring few-shot training on these many tasks, training on the Common Crawl~\cite{commonCrawl} dataset of roughly 1.0T words from the internet.

\subsection{CLIP}
The internet contains a wealth of image data accompanied by associated text, creating the opportunity for training paired vision and language models at scale. We use the particularly successful CLIP~\cite{clip} approach for our work, which has been trained on the WebImageText (WIT) dataset, containing 400M images from 500K language queries. For scalable performance, CLIP jointly trains a visual encoder and a Transformer language model to optimize a metric-learning objective. In other words, CLIP is trained to embed text and images into a common embedding space using a dataset of image/caption pairs, where the loss is minimized by increasing the cosine similarity between image-caption pairs that appear in the dataset while minimizing those of random image-caption pairings. Although the WIT dataset contains poor label accuracy compared to previous vision-specific datasets, its large size, when utilized by CLIP's modern VLM architecture, has produced a semantically aware visual understanding model which is state-of-the-art on many tasks.

\subsection{CA-SUM}
CA-SUM \cite{ca_sum} is a recently proposed unsupervised video summarization technique which we use as a baseline to compare our results against. This method tries to reason about the uniqueness and diversity of the video frames and presents a technique for a concentrated attention mechanism represented by a block diagonal sparse attention matrix. 

Given a set of video frames the method first converts these into feature representations using an off-the-shelf pre-trained model. These features form the inputs to a self-attention pipeline. The derived attention matrix $A$ is used as input for the concentrated attention mechanism. This mechanism produces a block diagonal sparse matrix $B$ that concentrates the information about the uniqueness and diversity of each frame within the block. The uniqueness $u_t$ for each frame is represented by the entropy over the entire frame sequence i.e. the entropy of each row of the attention matrix $A$. While the diversity $d_t$ is represented by the mean of the cosine similarity of each frame within this block to the ones that lie outside the block.

The final output of the network produces a set of frame-level scores that represents each frames' importance. At inference, given a temporal segmentation of the video (obtained using the KTS algorithm \cite{kts}), the importance of each segment is inferred by averaging the scores of the frames within this segment. Finally, by limiting the length of the video summary to be 15\% of the total length of video, the video summary is generated by solving the Knapsack problem.

\section{Method}
\begin{figure}
\includegraphics[width=0.48\textwidth]{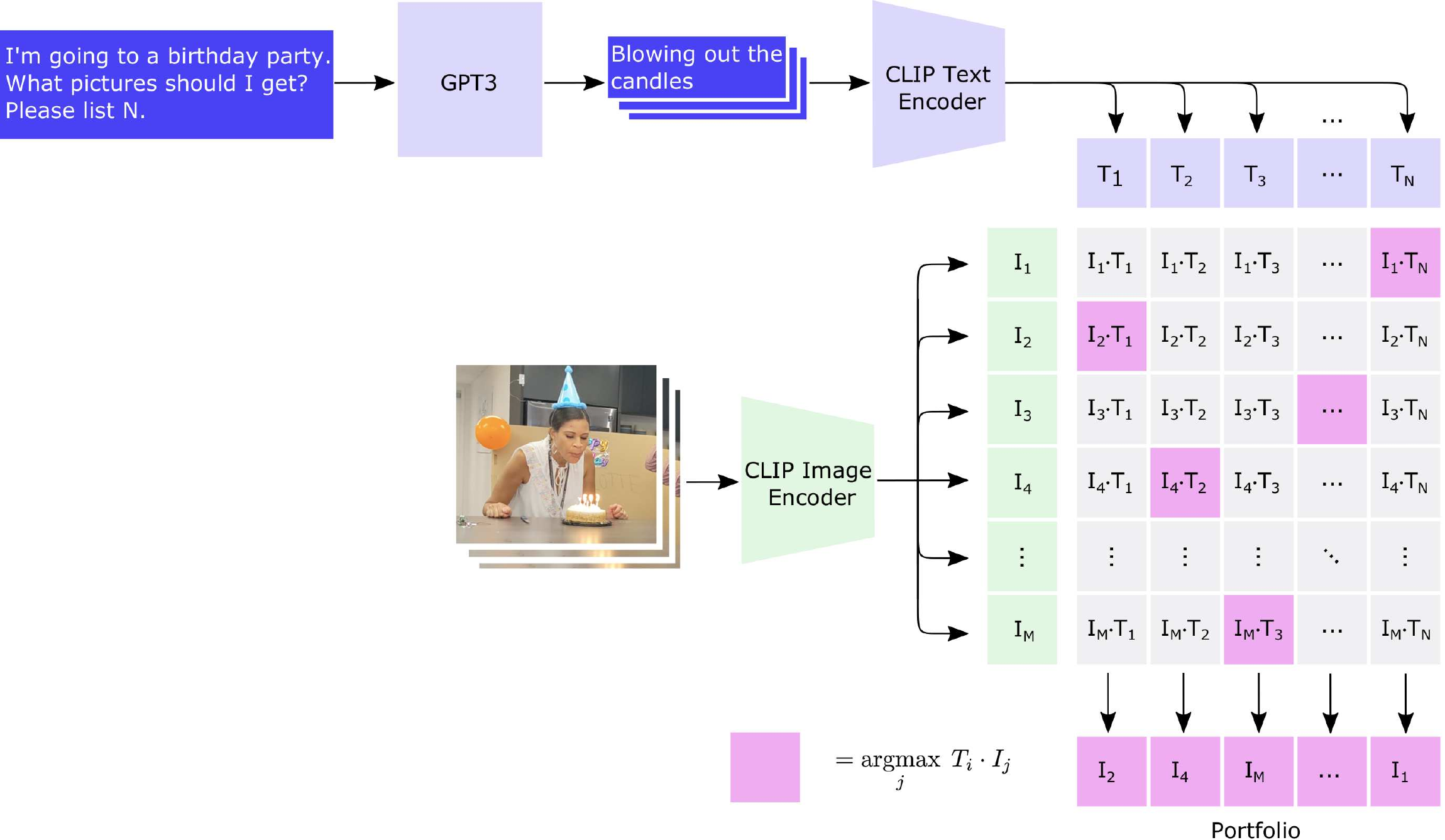}
\centering
\caption{Summary of our method. GPT3 is queried with a high level event description and asked to generate some phrases describing photographs that the photographer should try to capture. We then map these phrases and the images captured by the robot's camera into a common embedding space using CLIP. Next, we compute the pairwise dot product between all the phrase and and all the image embeddings. Finally, we apply the argmax operation over images for each phrase in order to find which image best describes each phrase. These images are output as the portfolio.}
\label{fig:architecture}
\end{figure}

Our approach is to leverage GPT3's world knowledge to extract textual task descriptions of the stereotypical photographs one would expect to see at an event given a prompt which describes the event at high level. An example of this approach in action is given in Figure \ref{fig:example}. In that example, the phrases returned by GPT3 include specific objects and concepts that can easily be identified in photographs (``person," ``cake," ``candles," "presents"). There are also activities which are often more challenging, but still possible, to recognize (``blowing out," ``opening"). Rather than parsing these concepts out explicitly and attempting to detect them, we simply embed the entire phrase output by GPT3 using CLIP's text encoder. Next, we also embed all the frames of the collected video using CLIP's image encoder. We then compute pairwise cosine similarities between each phrase embedding and each image embedding. Finally, we construct our portfolio by, for each GPT3-generated phrase, picking the image with the highest cosine similarity. This process is summarized in Figure \ref{fig:architecture}. The rest of this section describes some nuances of the implementation.

\subsection{Prompt engineering}
The term ``prompt engineering" refers to the design of prompts to the language models to get them to output reasonable results. In the case of our approach this is doubly relevant since we have two language-consuming models, GPT3 and CLIP.

\subsubsection{CLIP}
In prior work, researchers found that is was possible to use prompt engineering to employ CLIP as an Imagenet classifier. There, the authors found that prompt ensembling significantly increased classification performance. For example, instead of just comparing image embeddings to the text embeddings for ``a cat", they would average the embeddings for ``A painting of a cat," ``A cartoon cat," ``A realistic cat", etc. \cite{clip}. We experimented briefly with this kind of ensembling, using the same ensemble of modifiers as \cite{clip}, but did not find that it improved performance on our tasks. Instead, we directly fed the outputs of GPT3 into the CLIP text encoder.

We did find, however, that query phrases including descriptions of the composition (most commonly ``close up" and ``wide shot") tended to lead to particularly poor photo selections. As such, we rejected any phrase sets that GPT3 proposed that had any of the terms [close-up, closeup, close up, wide shot] within them. When a phrase set was rejected, another would be sampled. We also explicitly asked GPT3 to focus on content over composition (see Figure \ref{fig:BNF} for details).

\subsubsection{GPT3}
\begin{figure}[!]

\includegraphics[width=0.48\textwidth]{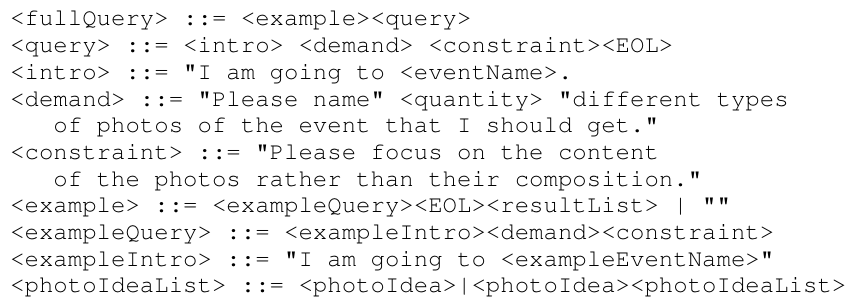}
\caption{BNF grammar for constructing GPT3 queries which create good ANSEL event plans. A photoIdea is an individual phrase such as ``A photo of the birthday cake". eventName is the name of the event you would like suggestions for, and exampleEventName is that for a different event that is used to construct the example. The same exampleEvent and photoIdeaList can be reused for priming across different values of eventName.}
\label{fig:BNF}
\end{figure}

We  use Backus–Naur form (BNF)~\cite{BNFjw1959c}
to facilitate and formalize the prompt engineering for our approach, but note that prompt engineering requires some degree of human intercession to optimize queries for different domains. For the event domains evaluated in this paper, our prompts have the form presented in in Figure \ref{fig:BNF}. This form creates a concrete specification for the desired outputs and allows for the priming of GPT3 using example input-output pairs, following the common pattern used in works such as \cite{flamingo}. 

Despite its apparent simplicity, obtaining good results can be elusive and depends on queries that are both precise enough to be effective, yet general enough to be understood by the language model and provide actionable results. We found that using the words ``photos of the event" rather than just ``photos" caused GPT3 to more reliably output the proper kinds of outputs. Without this, it would sometimes return different types of photos (landscape, HDR, black and white, etc). We also found that it was able to reliably generate lists of the requested length, and would always return responses in the enumerated format shown in Figure \ref{fig:example}, which made the outputs easy to parse. Without explicitly specifying the length of the list, the model tends to only output a few (2-3) types of photographs. In addition, based on the observation that CLIP often provides highly undesirable results when presented with directives related to composition, we ask the model to focus on content.

Finally, we observe that GPT3 behaves somewhat stochastically -- regenerating answers using the same prompt often leads to different phrasing. For example, it sometimes uses the term ``the birthday person" and other times it will be ``the birthday boy/girl."

\subsection{Dropping large embedding dimensions}
\begin{figure}
     \centering

     \begin{subfigure}[b]{0.25\textwidth}
         \centering
         \includegraphics[width=\textwidth]{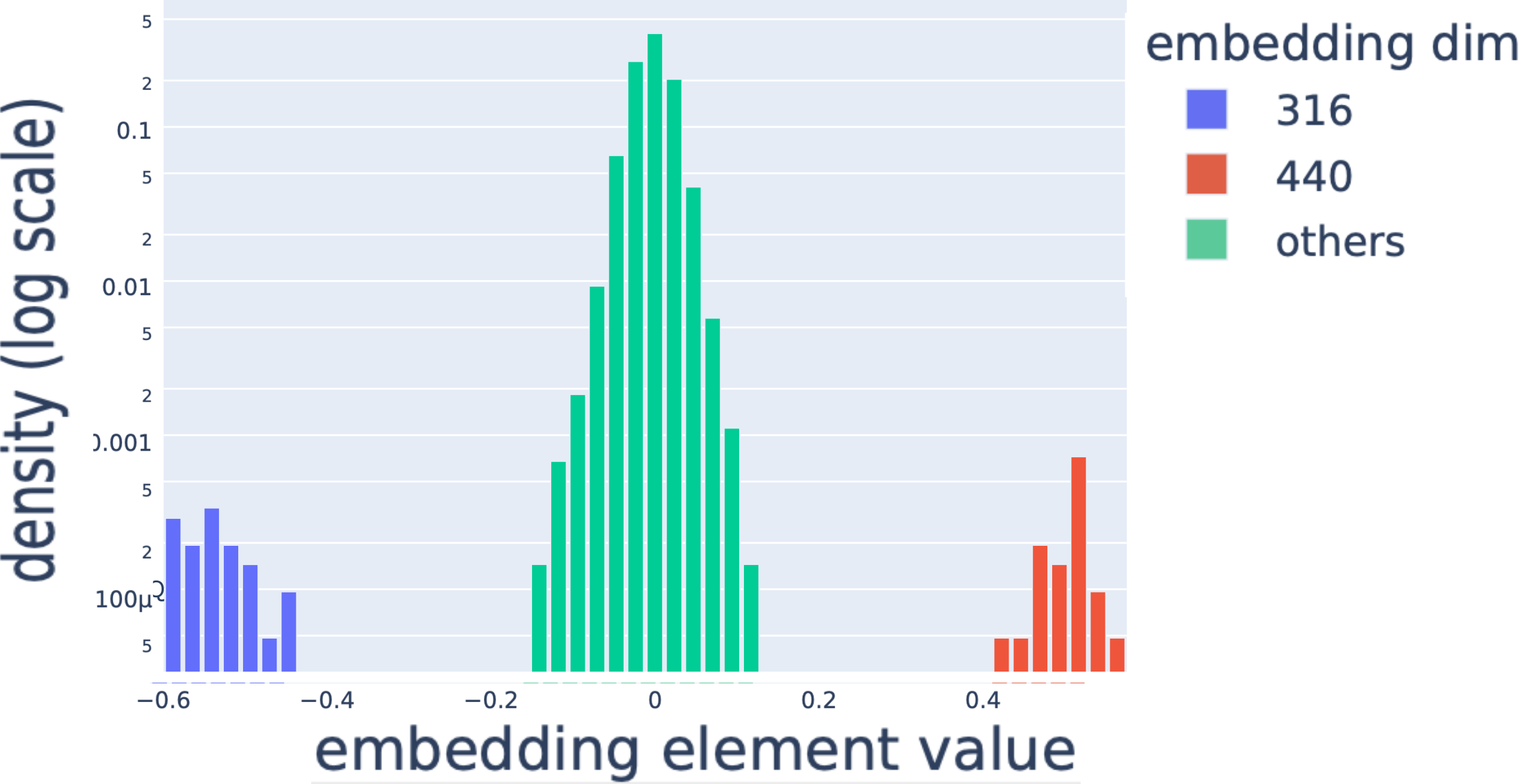}
         \caption{Histogram of text embedding element values over all captions.}
         \label{fig:text_embeds_histogram}
     \end{subfigure}
     \hfill
     \begin{subfigure}[b]{0.2\textwidth}
         \centering
         \includegraphics[width=\textwidth]{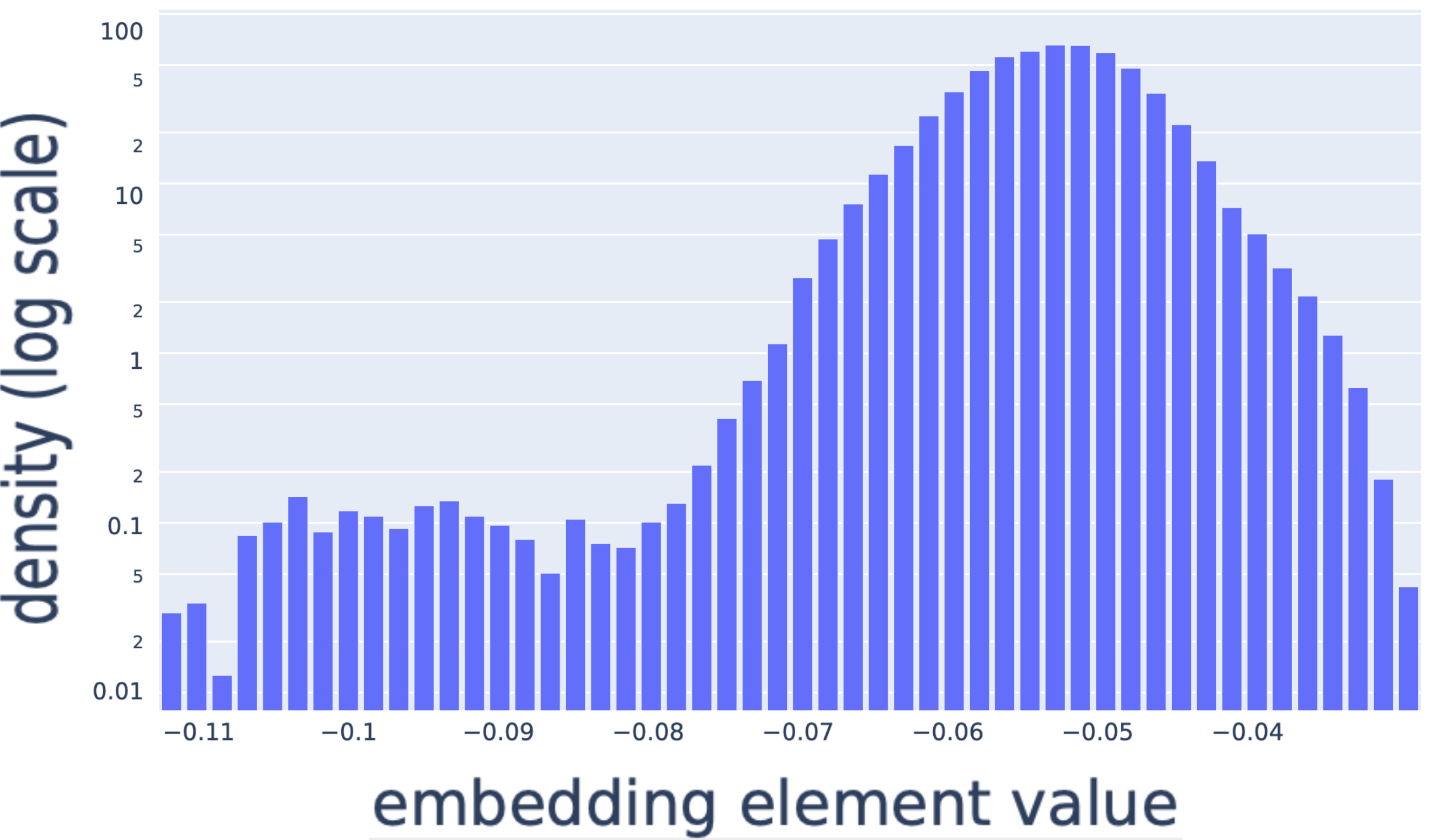}
         \caption{Histogram of image embedding element values at dimension 316.}
         \label{fig:img_embed_316}
     \end{subfigure}
     \hfill
     \begin{subfigure}[b]{0.2\textwidth}
         \centering
         \includegraphics[width=\textwidth]{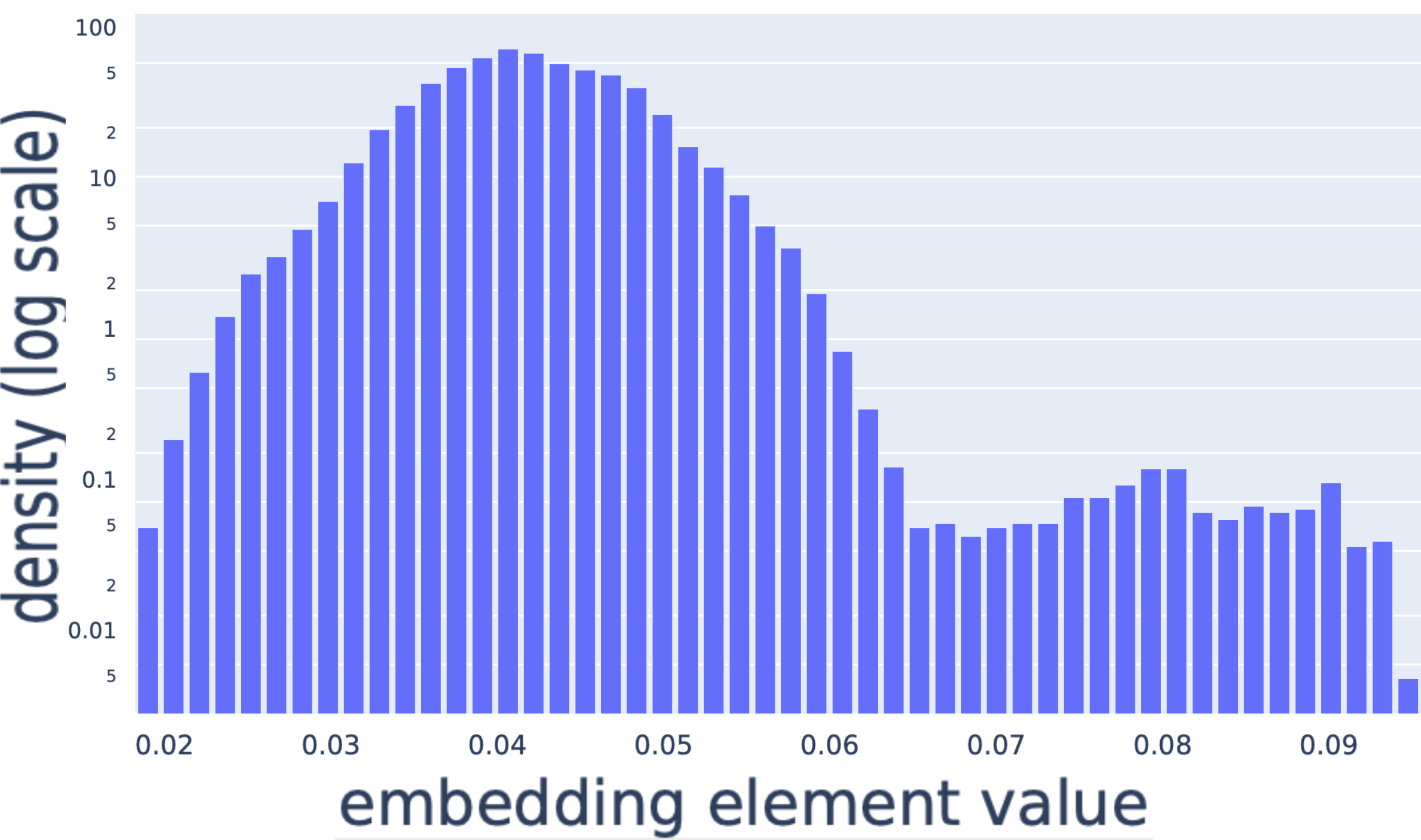}
         \caption{Histogram of image embedding element values at dimension 440.}
        \label{fig:img_embed_440}
     \end{subfigure}
    \begin{subfigure}[b]{0.2\textwidth}
         \centering
         \includegraphics[width=\textwidth]{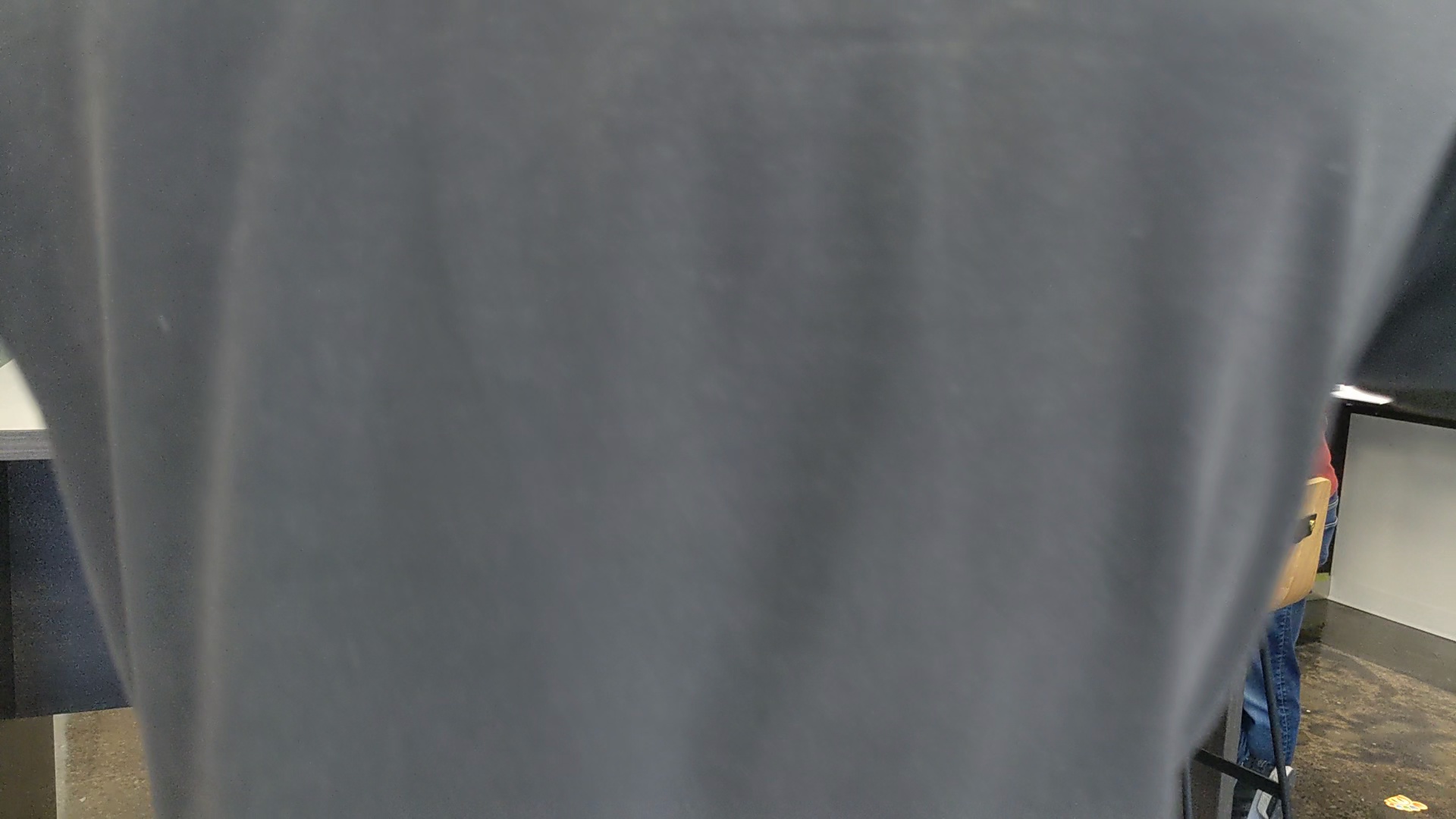}
         \caption{The image which has the largest magnitude at dimensions 316 and 440.}
         \label{fig:really_bad_image}
     \end{subfigure} 
        \caption{Two dimensions of the CLIP text embeddings have very large values, which tends to favor particularly bad images.}
        \label{fig:three graphs}
\end{figure}

During our experiments, we noticed that CLIP embeddings have two dimensions with values which are much larger than those of all the others. These are 316 and 440 for all text embeddings (Figure \ref{fig:text_embeds_histogram}), and 413 and 644 for all image embeddings (not shown). These large dimensions have a significant impact on the results; frames which have particularly large values on embedding dimensions 316 and 440 will have much higher cosine similarities to \textit{all} query phrases. Figures \ref{fig:img_embed_316} and \ref{fig:img_embed_440} show the distribution of the values of the image embedding dimensions 316 and 440 respectively. In both of these we observe a long tail of values that are significantly higher than the mean, meaning these frames should be selected particularly often. When we examine these frames with extremely large values of these two dimensions we find two interesting things. First, the same frame has the highest magnitude of both dimensions 316 and 440. Second, this frame, shown in Figure \ref{fig:really_bad_image}, is of extremely poor quality.

We have not seen these issues previously reported, and they are certainly not addressed in the original work which presented CLIP. We hypothesise that this quirk has not had a significant impact on the majority of other works using CLIP because they tend to work with more highly ``curated" datasets (like web data and youtube videos) than data coming from real-world robots, so they encounter fewer, or even none, ``garbage" frames like those in \ref{fig:really_bad_image}. We found that zeroing any dimensions with magnitude greater than 0.3 and then re-normalizing the embeddings was an effective method for reducing the dominance of these ``garbage" frames. 

\subsection{Faces}
\label{sec:face_cropping}
In view of our objective of finding satisfying photographs, we used human subjects and ``A/B" trials as the key objective assessment mechanism for our approach vis-a-vis alternative algorithmic approaches.   In  pre-existing automated photograph  acquisition systems,  although they are generally not driven by semantic criteria, let alone  general-purpose ones, attention to human faces is a common attribute. For  example, the Google Clips camera uses human faces to (sometimes) trigger video clip recording.  With this in mind, we crop 
the photographs we acquire around human faces that appear in the image.  If no human faces are detected, we filter out the image.

Faces are detected  using the Multi-Task Cascaded Convolutional Neural Networks (MTCNN) ~\cite{MTCNN2016} algorithm that detects faces using a cascade of h=three convolutional neural networks,  implemented using Tensorflow.  The bounding box for the ensemble of all faces is extracted, and expanded by  a small margin (10 {\em per cent}) and extended downward to the bottom of the image plane (size of detector  for the  extent of bodies is used).  The image is then zoomed and clipped around this bounding box to provide a more human-centric view of the scene.

\section{Evaluation}

\begin{figure}
\includegraphics[width=0.3\textwidth]{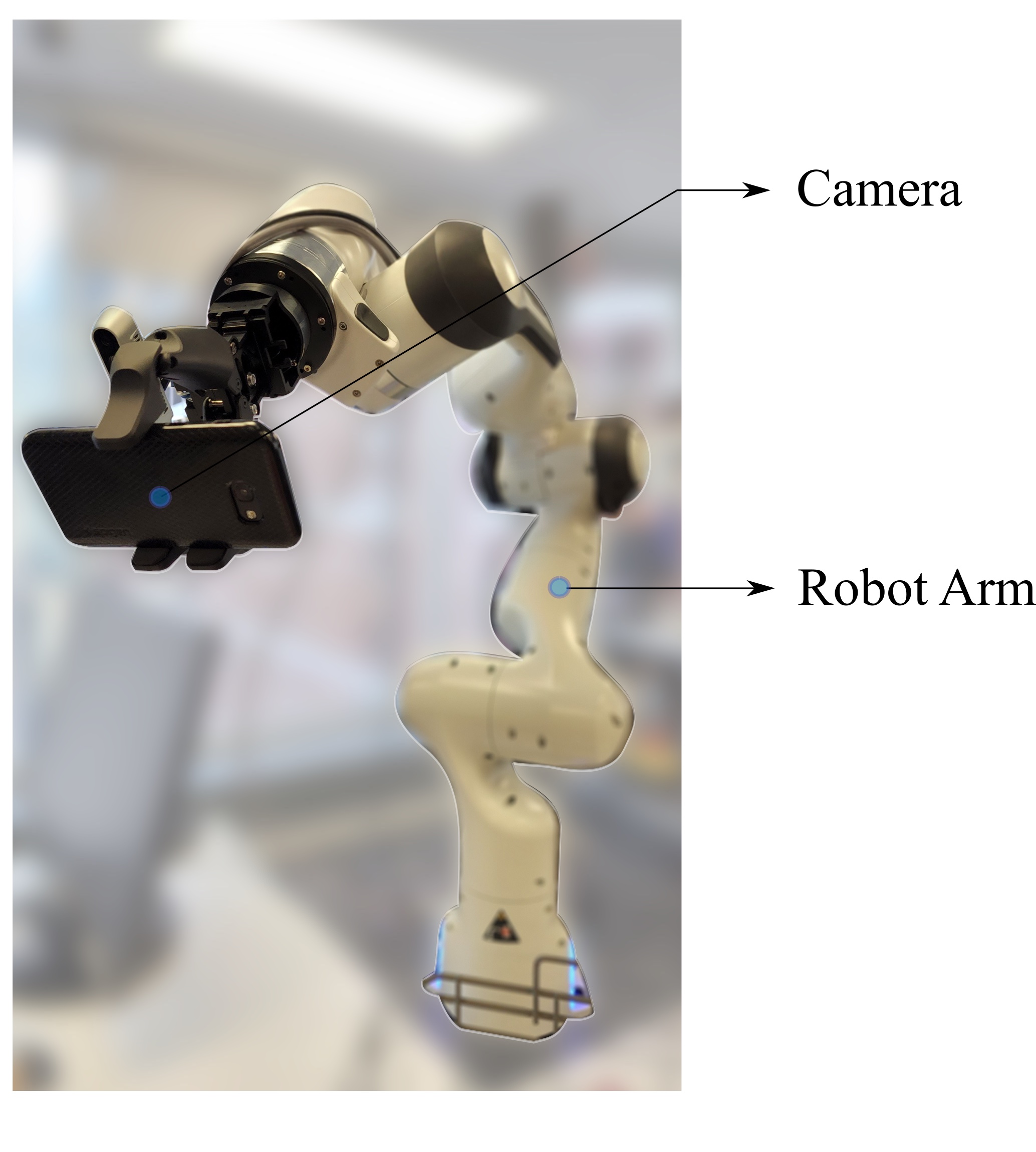}
\centering
\caption{ANSEL hardware embodiment: Robot arm holding the cell 
phone camera used to record data.}
\label{fig:arm}
\end{figure}

\begin{figure}[h]
\includegraphics[width=0.9\linewidth]{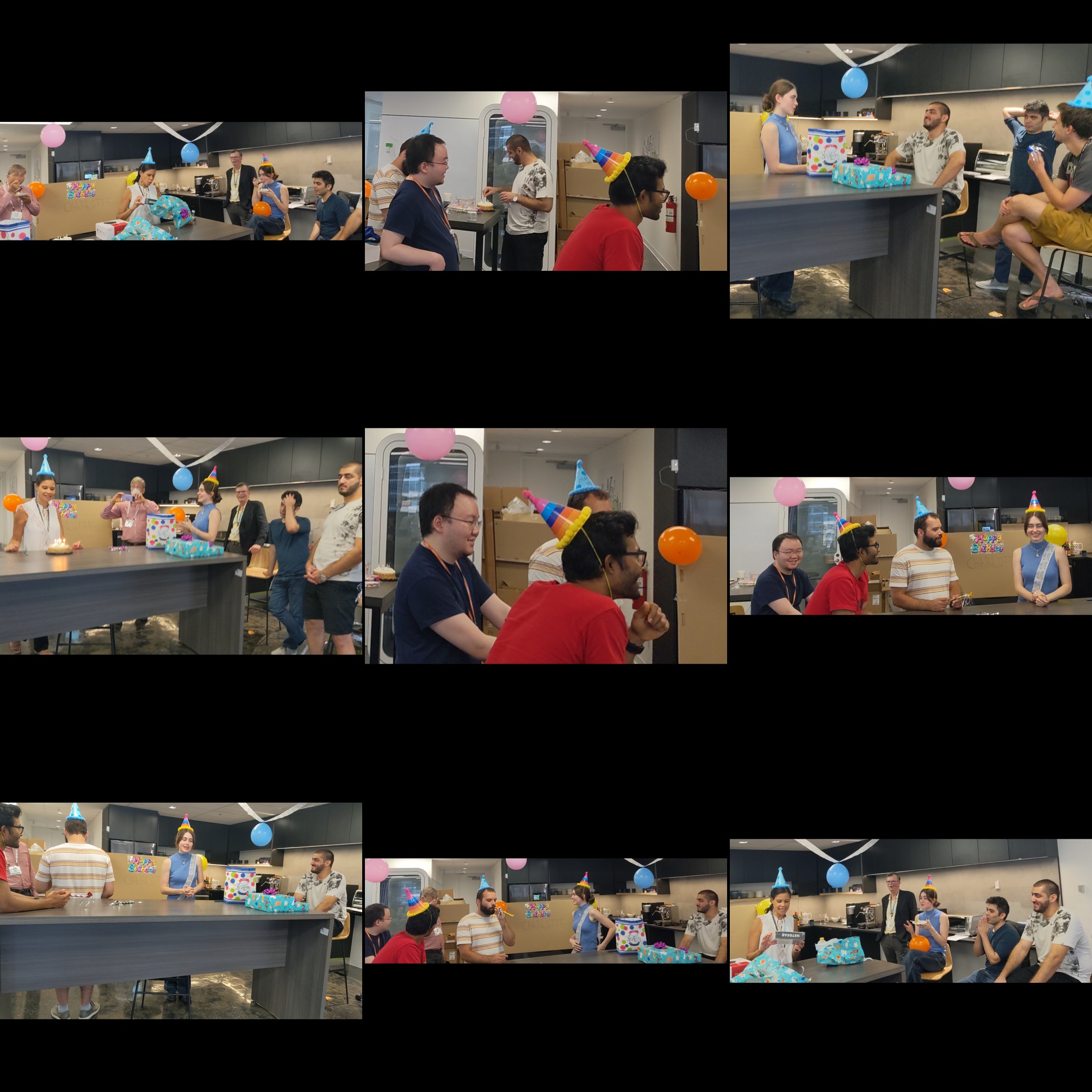}
\centering
\caption{Portfolio of photos from a birthday party scenario generated using ANSEL.}
\label{fig:partytile}
\end{figure}

To validate and evaluate our approach, we deployed our ANSEL Photobot system using a Franka Emika robot arm holding a cell phone used as a video recording device (shown in Fig.~\ref{fig:arm}).  Using an off-the-shelf cell phone camera allowed us to achieve high image quality and a degree of portability and hardware independence, but notably core processing operations are performed off board (ie. in ``the cloud'').

In this study, we had the robot collect video of 3 
simulated social events, a birthday party, a wine tasting event, 
and a painting class. These descriptions (``a birthday party", ``a wine tasting event", and ``a painting class") were used as eventNames in the query construction. Note that while we experimented briefly with the use of examples in priming GPT3, and preliminary results seemed promising, the results presented in this work omit the use of these priming examples.

The events were ``simulated'' in the sense that these were 
actual events, but they were convened for the purposes of this
paper in an office environment by people pretending to experience
social activities (although they did, in fact, experience the events in their full form including actually making paintings, eating cake, and learning about tastes). For each event, we generate a photographic summary of nine photos\footnote{CLIP hyperparameters: model=ViT-L/14@336px. GPT3 hyperparameters: model=text-davinci-002, temperature=0.7, max\_tokens=2000, top\_p=1, frequency\_penalty=0, presence\_penalty=0}, stitched together into a collage (an example is shown if Figure \ref{fig:partytile}, using both our method and a recent generic video summarization technique (CA-SUM) which serves as a baseline~\cite{ca_sum}.

We slightly adapted CA-SUM to match our problem setup. The original method selects short video clips (usually 0.5-2 seconds) and limits the total duration of these clips to 15\% of the original video length. We change this to limit the number of clips to 9, and selected the frame from the center of each clip. The model we used was trained on the TV-Sum dataset \cite{tvsum}. This dataset contains videos of similar length to our (15-30 minutes), along with user annotations which were used to perform model selection. We apply the same face-cropping procedure to the frames selected by CA-SUM as we do for our method.

We then surveyed 10 individuals to ascertain which collage was more semantically relevant to the each event type. We simultaneously presented them with collages for each method, and asked them to pick their preferred one. In order to prime the individuals to focus on semantics over composition, we first asked them to generate a list of nine photo types that would be appropriate to the event type before doing the evaluation. We also showed the users GPT3's suggestions after they had created their own and asked them to score each set of phrases (both their own and the ones from GPT3) on a scale of 0-10.

\section{Results}

\begin{figure}
\includegraphics[width=0.35\textwidth]{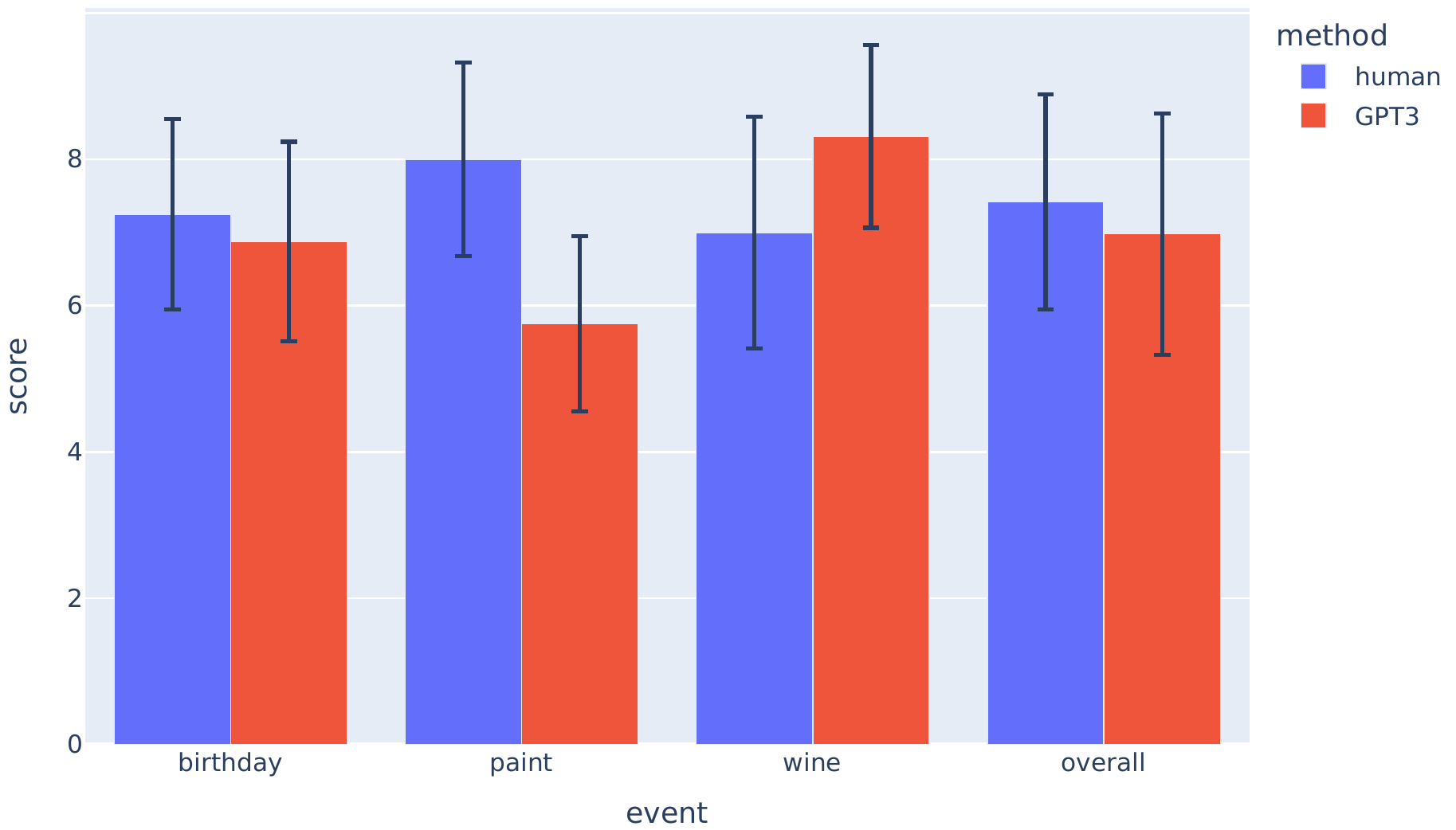}
\centering
\caption{Scores assigned by users to their own plan sets (human) and those generated by GPT3. Error bars represent standard deviation. Users were told to assign scores of 0-10 to each of the caption sets, so both systems were ranked as very good and also statistically indistinguishable.}
\label{fig:scores}
\end{figure}

All participants found the quality of the targets for photo subjects to be of high quality, often preferring them in retrospect over labels the subject themselves has suggested \footnote{Full list GPT3-generated suggestions for wine tasting event: People discussing the wine. The different types of wine on display. People mingling and networking.The venue of the event.The buffet or food that is being served.The decorations or theme of the event.The staff serving the wine.
The guests enjoying themselves}.  After seeing the GPT3 labels, subject's scoring of their own labels had an average value of 7.4/10  while their scores for the labels from GPT3 were 7.0/10 (see Figure \ref{fig:scores} for more details).   Clearly, GPT3 provided photographic directives at a high  semantic level that were as good as those generated by the humans, and in the opinion many subjects even better than their own.

\begin{figure}
\includegraphics[width=0.35\textwidth]{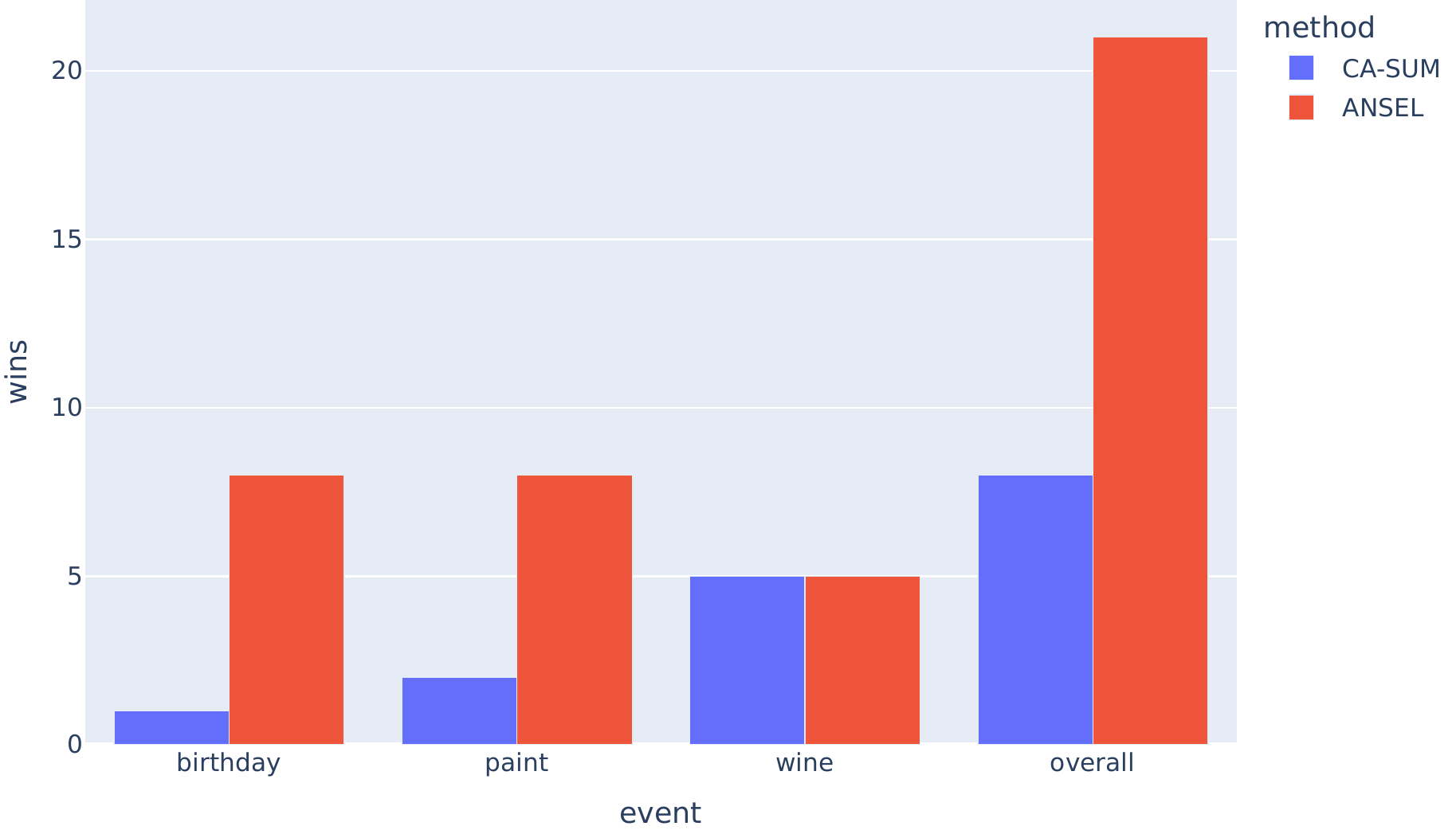}
\centering
\caption{Number of wins for ANSEL and CA-SUM on each of the events, as well as the aggregated value combining all events. The ANSEL algorithm is consistently preferred in 2/3 events, and ties in only one case.}
\label{fig:wins}
\end{figure}

ANSEL  outperforms CA-SUM in 2/3 events (birthday and paint), while tying it on the wine tasting, as illustrated by Figure \ref{fig:wins}. This confirms that ANSEL is capable of identifying the stereotypical, appropriate photographs associated with different types of social events. A potential explanation as to why ANSEL wins so decisively in two events and yet ties the third might be found in Figure \ref{fig:scores}. In the case of the wine tasting, on average people actually tended to prefer GPT3's captions to their own, which may indicate that they had weaker priors about the expected photos for this event type. Weaker expectations on the photo types could explain the apparent indifference to summarization method encountered in the wine event.

\section{Conclusion}
We have described an approach that uses an impressive level of implicit semantic knowledge embedded in a large language model to generate a robot planning systems that operates at a highly abstract level. We  introduced ANSEL Photobot, a robot photographer that can identify key events from a video stream using a combination of  LM and VLM models by adapting the semantic specifications to the event of interest.  Results show that our approach consistently generates photo portfolios that are consistently rated as more appropriate than video summarizing baselines by human evaluators. In the context of photographer/documentarian robotics specifically, we believe that there are opportunities to improve the quality of the photo portolios by incorporating image quality metrics as well (such as symmetry, affect, sharpness). These methods are beyond the scope of this paper as such enhancements would have made it more difficult to compare the results of the semantic filtering against baseline methods.

While it appears that GPT3 can produce near human level plans for the task examined in this paper, the challenge of grounding these concepts in the visual reality of robots remains significant. In Fig.~\ref{fig:example} for example,  it is clear that the subject is not ``the birthday person blowing out their candles." However, the person in the image is clearly blowing into a tube, and in fact it might be fair to say that this image matches the concept of ``blowing" much better than a person leaned over some candles. Furthermore, the person is wearing a party hat, and the words ``happy birthday" are clearly visible. It appears that CLIP is not capable of balancing this opulence of ``blowing-ness" and ``birthday-ness" against the dearth of candles. Assuming that VLMs continue to evolve as rapidly as they have in the last several years, we believe that these kinds of errors will soon become a thing of the past, and that LMs and VLMs will have a transformative impact on robotics.

\bibliographystyle{IEEEtran}
\bibliography{IEEEabrv,biblio}

\end{document}